# Unlearning Imperative: Securing Trustworthy and Responsible LLMs through Engineered Forgetting

James Jin Kang, Dang Bui, Thanh Pham and Huo-Chong Ling

RMIT University [702 Nguyen Van Linh, Tan Hung Ward, Ho Chi Minh City](...) Vietnam, +84 0909934001

Corresponding author: james.kang@rmit.edu.vn

**Abstract:** The growing use of large language models in sensitive domains has exposed a critical weakness: the inability to ensure that private information can be permanently forgotten. Yet these systems still lack reliable mechanisms to guarantee that sensitive information can be permanently removed once it has been used. Retraining from the beginning is prohibitively costly, and existing unlearning methods remain fragmented, difficult to verify, and often vulnerable to recovery.

This paper surveys recent research on machine unlearning for LLMs and considers how far current approaches can address these challenges. We review methods for evaluating whether forgetting has occurred, the resilience of unlearned models against adversarial attacks, and mechanisms that can support user trust when model complexity or proprietary limits restrict transparency. Technical solutions such as differential privacy, homomorphic encryption, federated learning, and ephemeral memory are examined alongside institutional safeguards including auditing practices and regulatory frameworks.

The review finds steady progress, but robust and verifiable unlearning is still unresolved. Efficient techniques that avoid costly retraining, stronger defenses against adversarial recovery, and governance structures that reinforce accountability are needed if LLMs are to be deployed safely in sensitive applications. By integrating technical and organizational perspectives, this study outlines a pathway toward AI systems that can be required to forget, while maintaining both privacy and public trust.

*Keywords*: Data Privacy, Large Language Models (LLMs), Ethical AI, Adversarial Manipulation, Regulatory Compliance, Machine Unlearning, User Trust

## I. Introduction

For artificial intelligence to be truly trustworthy, Large Language Models (LLMs) must not only learn effectively but also be able to forget responsibly [1]. LLMs have quickly become vital in fields that handle sensitive information, such as healthcare, defense, and public services [2]. As these models grow smarter and more widely used, concerns about privacy and control are becoming harder to ignore. Unlike traditional systems where data can be easily tracked and deleted, erasing information from an AI model is not like deleting a folder on your computer [3]. Once trained, the model stores knowledge in complex and often hidden ways, making it difficult to remove specific pieces of data. This creates risks of unintended exposure and raises serious questions about accountability [4].

One promising approach to these challenges is machine unlearning, the ability to selectively erase certain data from a model after training [5]. While this concept offers hope for stronger privacy and control, it also opens new technical and practical questions. How do we reliably confirm that data has truly been removed? How can models be protected against attempts to recover deleted information? And how can we maintain user trust when the inner workings of these models remain largely hidden? This section explores these issues in detail, laying out the key challenges and opportunities involved in unlearning a practical, trustworthy part of AI systems.

### A. User Trust in LLMs

Trust in LLMs is about more than whether they give the right answer to the user. People also want to know that their personal or sensitive data is safe, that the model won't leak private details to others, and that it can withstand attacks designed to exploit weaknesses [6]. Privacy and security are therefore at the heart of building confidence in these systems [7]. Yet even with strong protections, there is still the problem that sensitive or unwanted information can remain "soaked or locked in" the model once it has been trained [8]. To deal with this, there should be a way to reliably remove or undo what should not be there. This is where machine unlearning comes in [5]. By giving users and regulators confidence that data can be erased when necessary, unlearning helps turn abstract promises of trust into something concrete. For this reason, the third research question asks how unlearning can support accountability and smooth deployment of LLMs to provide trust.

### B. Machine Unlearning and Its Importance

The rapid integration of generative artificial intelligence (GenAI) and LLMs into decision making systems marks a turning point in both technical capability and societal impact [9]. From assisting doctors with administrative tasks to supporting legal review, defense analysis, and public service delivery, these models are now applied to domains that rely on sensitive and personal data [10]. Their adoption reflects growing confidence in their utility, yet it also magnifies the consequences of failure. When models leak, retain, or misuse private information, the risks extend beyond algorithmic error to matters of legal compliance, institutional trust, and ethical responsibility [11].

Unlike traditional databases, LLMs were not designed with privacy at their core [12]. Once data enters the training pipeline or fine-tuning stage, they may become diffusely encoded in model [3]. Even if the original files are deleted, fragments of personal information may remain recoverable through indirect queries or adversarial prompting. This structural complexity raises a fundamental question: how can users, regulators, and developers ensure that AI systems do not indefinitely retain or expose data they were never meant to hold? Figure 1 illustrates the process of unlearning in machine learning. The target data, marked in red, represents the information to be deleted from the trained model. Through the unlearning algorithm, the model is adjusted so that the influence of the target data is effectively removed. In the relearning stage, the target data is replaced with black nodes, showing that the model no longer retains the deleted information while maintaining comparable performance to a retrained model [12].

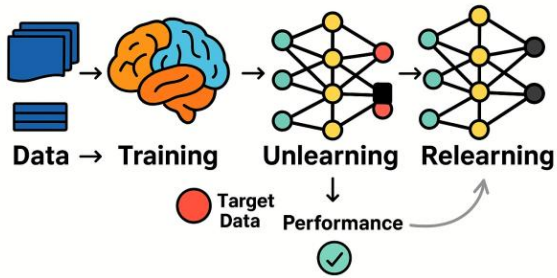

Fig. 1: Illustration of the machine unlearning process, where target data are from the trained model and replaced during relearning to approximate the performance of a retrained model.

### C. Emergence and Urgency of Machine Unlearning

Machine unlearning has emerged as a promising response to the challenges of data privacy in large language models (LLMs). Unlike retraining a model from scratch, unlearning techniques aim to selectively remove the influence of specific data while preserving overall performance. This is not simply a matter of deleting files, but of erasing traces of information encoded within model parameters so that the data no longer shapes outputs [13]. This matters because under regulations such as the European Union's General Data Protection Regulation (GDPR), individuals have a legal right to request deletion of their personal data, often known as the "right to be forgotten" Yet LLMs are designed to remember, making forgetting technically difficult [14]. Unlearning offers a potential compromise: by modifying only the relevant parts of a model, organizations may comply with privacy requests more efficiently, avoiding the prohibitive costs of retraining. However, significant challenges remain. Once deletion is claimed, how can we confirm that the information is truly gone? What proof can be offered to users, auditors, or regulators that the model no longer encodes or reveals sensitive data [15]? Evidence shows that even after apparent removal, LLMs can unintentionally reproduce memorized details such as names, addresses, or medical notes, creating serious risks in high-stakes domains like healthcare, defense, or law [16], [17]. Retraining without the sensitive data is the default solution, but it is costly and impractical, large LLMs may require millions of dollars in computation and weeks of effort [18]. To address this, researchers at the University of California, Riverside have introduced a "source-free certified unlearning" method that allows models to erase private or copyrighted data without access to the original training datasets [19]. This innovation significantly reduces computational burden while maintaining functionality, offering a more practical pathway to trustworthy and compliant AI systems. Other approaches, such as ephemeral memory design (where user data is processed in temporary environments and not written into weights) or external auditing mechanisms, provide partial solutions [20]. Yet these remain experimental, depend on third-party trust, or fail to tackle the deeper problem of data embedded in model parameters. Moreover, adversarial prompting shows that even "cleaned" models may regurgitate deleted information under certain conditions [3].

### D. Research Gap

At present, there is no widely accepted standard for measuring, verifying, or guaranteeing unlearning in LLMs [10]. Efforts are fragmented across research groups, startups, and large AI labs, and progress remains uneven. In effect, while AI capabilities have advanced rapidly, investments in privacy assurance, accountability, and trust have lagged.

This creates a pressing research gap: how can machine unlearning be made reliable, efficient, and verifiable at scale? Addressing this question is vital if LLMs are to be deployed responsibly in domains where mistakes or breaches could have serious human, legal, and societal consequences.

### E. Objectives and Research Questions

This review addresses these challenges by surveying the state of machine unlearning for LLMs and evaluating both technical approaches and broader governance mechanisms. Specifically, the paper is guided by three research questions (RQs):

**RQ1**: How can the effective removal of sensitive data from LLMs be measured and validated without excessive computational costs?

**RQ2**: What technical safeguards can protect unlearned LLMs against adversarial recovery and unintended data leakage?

**RQ3**: How can user trust in LLMs be enhanced when transparency is constrained by model complexity or proprietary restrictions?

Having addressed the technical feasibility (RQ1) and resilience against adversarial threats (RQ2), RQ3 turns to the critical question of how these capabilities translate into user trust under real-world constraints. By addressing these questions, this paper clarifies the current landscape of unlearning research, highlights limitations, and identifies opportunities for progress

## II. Methodology

Our study followed a structured literature review to select works centered on the three research questions. To ensure relevance, we prioritized databases with strong coverage of machine learning (ML), natural language processing (NLP), security and privacy, and applied trust domain research.

Within this scope, a substantial portion of the reviewed works originate from arXiv, which has become the primary distribution for cutting-edge LLM unlearning research as many works first appear as preprints. The second major cluster is drawn from the ACL Anthology, covering conference and workshop proceedings from ACL, EMNLP, NAACL, and related venues. Additional contributions come from NeurIPS and its affiliated workshops, as well as ICML/ICLR. In particular, NeurIPS workshops provide specialized forums on LLM unlearning and robustness, often presenting exploratory studies and benchmarks not available in the main proceedings. Complementary works are taken from IEEE/ACM journals and conferences, along with peer-reviewed journals such as Artificial Intelligence Review, Knowledge-Based Systems, JMIR and NPJ Digital Medicine. Finally, regulatory and standards documents, including the General Data Protection Regulation [21], the EU Artificial Intelligence Act [22], and the NIST AI Risk Management Framework [23], were included to provide legal and governance context. In defining our corpus, we considered but ultimately excluded broader indexing services such as Scopus and Web of Science, as these provide limited coverage of recent preprints and LLM-specific security research.

For each source, two search queries were applied. The primary query targeted unlearning, forgetting, privacy-preserving and trust mechanisms in LLMs using the tailored Boolean queries as follows:

> ("machine unlearning" OR "knowledge unlearning" OR "data forgetting" OR "digital forgetting") AND ("large language model" OR "LLM" OR "transformer") AND ("privacy" OR "data security" OR "data leak" OR "adversarial attack" OR "robustness" OR "trust")

The secondary query extended coverage to essential adjacent works that may not explicitly use the keyword "unlearning", but are critical for RQ2 security evaluation. This is because many relevant works on security and adversarial robustness do not mention unlearning. Therefore, the secondary query contains the following tailored Boolean queries:

> ("adversarial training" OR "membership inference" OR "data extraction") AND ("large language model" OR "LLM" OR "transformer")

The core search spanned January 2022 to July 2025, reflecting the rapid acceleration of LLM-specific unlearning research during this period. This time frame aligns with the release of large-scale generative models such as GPT-3 and ChatGPT, which triggered a surge of work on model safety, privacy and controllability. This search was supplemented with foundational search tracing research prior to 2022 that introduced core studies on adversarial threat models, membership inference, privacy-preserving mechanisms and early interpretability methods. Although not LLM-specific, these works provide methodological and evaluative frameworks that continue to guide the design of unlearning strategies.

Publications were screened on two stages. The title and abstract screening excluded works that were clearly outside the scope, followed by full-text screening to assess methodological relevance. The process was iterative, with refinement through cross-checking references from key papers to ensure coverage of work relevant to the research questions. Inclusion criteria required studies to address at least one of the following: methods for evaluating or verifying unlearning in LLMs; adversarial or data leakage threats and corresponding safeguards; or mechanisms for fostering user trust in LLMs. Peer-reviewed and preprint publications were considered to capture the most recent developments, while pre-2022 studies were included for foundational threat models and defense mechanisms. Exclusion criteria removed studies that lacked a direct connection to unlearning, adversarial robustness, privacy or trust in LLMs, as well as those limited to high-level commentary without substantial methodology contributions.

Although comprehensive, the methodology has several limitations. First, the search was restricted to English language publications, and hence relevant work in other languages was excluded. Second, the inclusion of preprints introduces the possibility of including studies that have not undergone peer review, but this inclusion was necessary to capture the most current advances in this rapidly evolving field.

The final reference set was organized according to three guiding research questions: RQ1- Unlearning evaluation and verification [24]-[50], RQ2 - Towards robust unlearning for LLM [51]-[68], and RQ3 - Promoting User Trust in LLMs [21]-[23], [69]-[90]. These domains collectively provide the foundation for the literature review that follows. Within each domain, the included works were further classified by publication type, namely journal articles, conference proceedings, preprints, and regulations and standards. Journals and major conference proceedings represent established peer-reviewed contributions, while preprints highlight emerging work that has not yet undergone formal peer review but is important for better representation of the research questions. References to regulations and standards were included as authoritative sources relevant to governance and policy. For consistency, workshops and findings tracks were grouped under conference proceedings. This classification strengthens the transparency and reliability of the reviewed studies by clarifying the balance between peer-reviewed and preliminary contributions. Figure 2 illustrates the distribution of references by publication type.

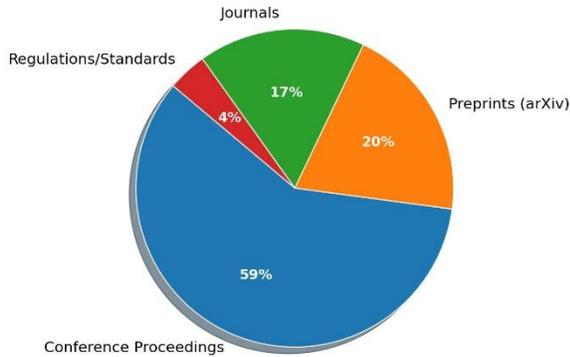

Fig. 2: Distribution of references by publication type (n=70).

The majority of publications, totaling 41 articles (59% of the total), appeared in conference proceedings, followed by 14 preprints (20%), 12 journal articles (17%), and 3 regulations and standards (4%). This distribution indicates that the research area remains active and evolving, with a growing but not yet dominant body of peer-reviewed journal contributions. Additionally, it allows readers to understand the balance between peer-reviewed and non-peer-reviewed sources.

## III. LITERATURE REVIEW

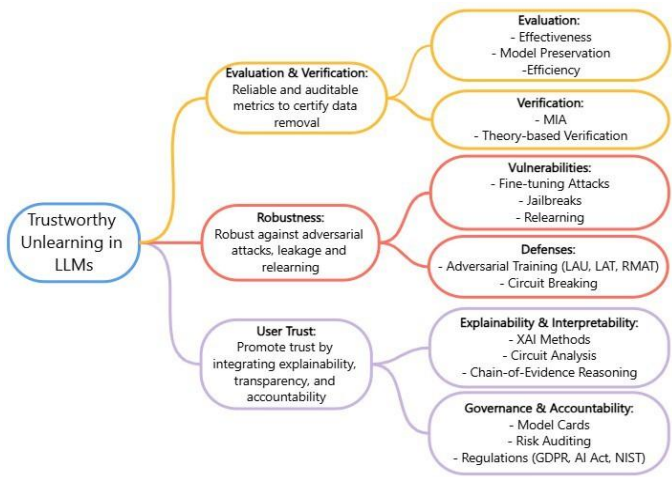

Fig. 3: Mind-map of the literature review framework for trustworthy unlearning in LLMs, structured around three research questions: evaluation and verification (RQ1), robustness (RQ2), and user trust (RQ3).

As illustrated in Figure 3, the literature review that follows is structured around three central goals: (1) identifying approaches to evaluate whether data has been effectively removed without requiring extensive computational resources, (2) investigating security mechanisms that prevent erased information from resurfacing or leaking, and (3) exploring strategies to foster user trust in model outputs despite the opacity of the underlying architectures.

### A. Unlearning Evaluation and Verification

While current safeguards aim to secure data before, during, and after training, challenges remain with leakage. Language models internalize data in ways that are opaque and hard to reverse; even if source files are deleted, fragments may still be recovered through indirect queries or adversarial prompts. This creates a need for technical solutions that can remove sensitive information without degrading performance.

Machine unlearning offers such an approach. Rather than retraining a model from scratch, it seeks to remove specified data and its influence by targeting traces within model parameters so that the information no longer shapes outputs. The growing deployment of LLMs in real-world applications that handle personal, sensitive, and legally regulated information has sparked increasing research interest in LLMs unlearning.

As such, a central challenge in this field is assessing whether the unlearning process has successfully and irreversibly erased the unwanted data [24], [25]. This challenge arises from the generative nature of LLMs, as well as the large scale and complexity of the underlying architecture, making it difficult to trace and verify the removal of embedded information. In this section, we explore both evaluation and verification methods for LLM unlearning. While these concepts overlap, their primary difference lies in their focus: evaluation measures the performance of methods against the established objectives, whereas verification aims to certify unlearning and offer a bounded guarantee [20]. The summary of the evaluation metrics and verification methods for LLMs unlearning is presented in Table I.

*1) Unlearning Evaluation:* To understand the evaluation of unlearning, it is essential to first unpack the core objectives of LLMs unlearning. Formally, these objectives include effectiveness, model preservation, and efficiency [10], [15], [26]. In essence, the goal of unlearning is to effectively and efficiently remove the influence of specific unwanted data without degrading the model performance on the retain data. Evaluation metrics provide model providers with theoretical criteria to assess the technical success of unlearning algorithms.

*a) Effectiveness:* Unlearning effectiveness refers to the extent to which the unlearned model behaves as if it were never trained on forgotten data. This concept is also termed forget quality or unlearning quality in literature.

- **Performance on the forget set:** Unlearning effectiveness is commonly evaluated using accuracy on the forget set, as demonstrated in various studies [27]-[34]. This metric reflects the degree to which the model retains knowledge of the erased data. Post unlearning, the model should exhibit significantly reduced performance on the forget set, indicating that it no longer leverages information from the forgotten data during inference. Ideally, the accuracy metric should closely resemble that of a model trained without forget samples, or degrade to the level of random guessing [28].

- **Truth Ratio:** Introduced in the TOFU unlearning benchmark dataset [31] and applied in subsequent studies [32]-[36], the Truth Ratio is computed on the forget set for both the reference model (trained without the forget set) and the unlearned model. The resulting distributions are compared using the Kolmogorov-Smirnov (KS) test. A high p-value indicates that the two models behave similarly on the forget set, suggesting successful unlearning. However, this metric is only applicable for theoretical evaluation of unlearning frameworks where the target LLMs were fine-tuned on TOFU before unlearning, since the benchmark relies on the injection of fictitious data that did not exist within the original training set.
- **Similarity Metrics:** Similarity-based evaluation, using metrics such as BLEU and ROUGE-L, has been utilized to assess the effectiveness of unlearning methods [32]-[34], [37]. These are typically sentence-level and token-level similarity metrics, which measure lexical, or n-gram overlaps between the outputs of the unlearned model and the original model on the forgot sample. A high similarity score indicates insufficient unlearning, suggesting that the model still retains knowledge of the erased data. However, this approach is more effective for utility preservation assessment, where high similarity is desirable. In contrast, for unlearning effectiveness evaluation, low similarity does not necessarily guarantee successful forgetting, since LLMs are generative and can reproduce forgotten knowledge in paraphrased or semantically equivalent forms.
- **Membership Inference Attacks (MIA)**: MIA is an attack technique that aims to determine whether a specific data point was part of a model's training data [38]. In the context of LLMs unlearning, MIA success rate is used to evaluate unlearning effectiveness by testing whether certain data has been successfully removed from the model, and the behaviors of unlearned data are not exhibited in the unlearned models' outputs. Due to its applicability, especially when retraining for evaluation is impractical, MIA has become a standard evaluation method in recent studies [27]-[29], [32], [39], [40].

*b) Model Preservation:* Utility preservation is a key requirement in unlearning, aiming to ensure that the unlearned model maintains its performance on non-target data and general downstream tasks as compared to the original model.

- **Performance on retain data:** Accuracy on the retain set is applied to assess model preservation in various works [28]-[34], similar to its role in evaluating unlearning effectiveness. If the unlearned model achieves comparable accuracy on the retain set between the unlearned and the original model, it is indicated that the unlearning process has not affected the model's utility on knowledge that is outside of the target scope.
- **Similarity Metrics:** BLEU and ROUGE-L are widely adopted for assessing the preservation of model capabilities [30]-[34], [37]. These metrics compare the responses of the unlearned model to those of the original model on non-target data, under the assumption that utility is preserved when the outputs remain consistent before and after unlearning. A high similarity score indicates that the model's behavior on retain data has been maintained. Compared to their application in unlearning effectiveness evaluation, similarity metrics offer a more coherent and empirically grounded basis for assessing model utility.
- **Performance on general knowledge benchmarks:** To evaluate the overall capabilities post-unlearning, many studies employ the general language and knowledge benchmarks to assess model performance on downstream tasks [28], [30], [41]-[43]. The Massive Multitask Language Understanding (MMLU) [44] and TruthfulQA [45] benchmarks are among the most widely used, covering a broad range of topics across natural and social sciences. These benchmarks are designed to assess LLMs' proficiency in factual knowledge, reasoning, and language comprehension.

*c) Efficiency:* Reducing computational overhead is the main motivation behind unlearning since retraining the model from scratch offers guaranteed data removal yet remains expensive and impractical for massive architectures like LLMs. Most studies demonstrate efficiency via computational time [27], [28], [37], [40], often with the capped limit of 20% of the retraining time to be considered "efficient" [46]. Memory cost is also considered and evaluated in various studies [27], [40].

*2) Unlearning Verification:* While evaluation primarily assesses unlearning from a technical perspective, verification shifts the focus toward the rights and assurances of data contributors. Unlearning verification aims to provide formal guarantees and verifiable evidence that the model has genuinely forgotten specific data [20], [47], enabling individuals and organizations to trust that their information no longer influences the model's behavior.

- **MIA:** Besides its role in evaluation, MIA can also be leveraged as a post-hoc verification tool for unlearning, as suggested by [20], [47], [48]. By detecting whether a specific data point belongs to the training data, MIA is effective at identifying knowledge leaks and providing certification for unlearning effectiveness. Du et al. [49] also proposed the Unlearning Likelihood Ratio Attack+ (U-LiRA+) specifically for unlearning audit, which can detect the membership information of unlearned textual data with high confidence.
- **Theory-based Verification:** Some LLMs unlearning algorithms are designed with privacy protection

TABLE I: Summary of evaluation metrics and verification methods for LLM unlearning algorithms.

| Category | Objective | Metric/Method | Summary | References |
|---|---|---|---|---|
| **Evaluation** | Effectiveness | Performance on Forget Set | Measures accuracy drops on the forget set; ideally approaches random guessing or matches the accuracy of the model trained without the target data. | [27]-[34] |
| | | Truth Ratio | KS-test on model responses on forget samples of the unlearned and the reference model; a high p-value suggests successful unlearning. | [31]-[36] |
| | | Similarity Metrics | Computes the similarity score of the unlearned and the original model on forget samples; a high similarity implies ineffective unlearning. | [32]-[34], [37] |
| | | MIA Success Rate | Assess whether the unlearned data is still memorised; a high attack success rate post-unlearning indicates insufficient unlearning. | [27]-[29], [32], [38]-[40] |
| | Model Preservation | Performance on Retain Set | Measures the accuracy on non-target data to assess utility preservation; a high accuracy is desirable. | [28]-[34] |
| | | Similarity Metrics | Computes the similarity score of the unlearned and the original model on retain samples; a high similarity implies model utility preservation on non-target data. | [30]-[34], [37] |
| | | Performance on General Knowledge Benchmarks | Measures the model general capabilities in factual knowledge, reasoning, and language comprehension post-unlearning. | [28], [30], [41]-[45] |
| | Efficiency | Computational Time | Measures runtime and GPU requirements compared to retraining. | [27], [28], [37], [40] |
| | | Memory Cost | Measures memory usage during the unlearning process. | [27], [40] |
| **Verification** | Effectiveness | MIA | Post-hoc auditing tool for verifying data forgetting. | [20], [47]-[49] |
| | | Theory-based Verification | Offers a theoretical guarantee of data removal via privacy protection mechanism in design properties. | [34], [50] |

mechanisms and offer formal verification guarantee. Madmud et al. [34] introduced the differential privacy-based unlearning framework DP2Unlearning, which provides verifiable forgetting guarantees. Additionally, Zuo et al. [50] proposed the Federated TrustChain framework, which integrates blockchain technology with federated learning to support LLMs unlearning. The application of blockchain ensures all unlearning operations are provable, secure, and transparent.

*3) Detailed Evaluation and Validation Metrics:* Building on the broad categories of effectiveness, preservation, and efficiency introduced above, recent works have proposed fine-grained evaluation and validation techniques for LLM unlearning. These methods aim to capture subtle forms of residual memorization, improve robustness against adversarial probing, and balance forgetting with utility preservation.

The detailed findings can be found in Table II. At the forget set level, metrics such as LiRA, KL-divergence, Truth Ratio, ROUGE-L, and probability bounds have been widely applied [27]-[29], [35], [36], [39], [40]. Entropy-based bounds demonstrate higher sensitivity to residual leakage than deterministic tests [28], [35], while teacher-student distillation strategies achieve strong forget quality with stable retention [36]. Forgetting set accuracy remains the primary indicator of successful unlearning, ideally dropping to random guessing or reference-model behavior [27], [34]. Complementary measures such as the Truth Ratio [31]-[36] and similarity scores [32]-[34], [37] further refine analysis, though adversarial designed probes can still reveal hidden memorization even when conventional metrics suggest success [38]-[40].

Leakage resilience is also tested under adversarial conditions, particularly membership inference attacks (MIAs) and exposure-based auditing [27], [29], [32], [38]-[40]. Empirical results show that DP-SGD and SISA retraining reduce MIA success rates [27], [32], while auditing methods such as ULIRA and LiRA-Forget effectively expose residual memorization [39], [40].

Verification methods extend beyond empirical evaluation, offering stronger assurances of forgetting. Post-hoc auditing via LiRA, ULIRA, KS-tests, and Truth Ratio remains reactive, identifying leakage only after deployment [20], [47]-[49]. Theory-based methods such as DP$^2$Unlearning [34] and Federated TrustChain [50] provide formal guarantees, though they often rely on restrictive assumptions that limit applicability to large-scale LLMs. Hybrid approaches combining empirical auditing with theoretical bounds provide more balanced assurance but still face scalability challenges. Model preservation is typically assessed using retain set accuracy, divergence measures such as JSD and harmonic mean utility scores, and large-scale benchmarks including MMLU and AGIEval [28], [30], [41]-[45]. Structured methods such as Split-Unlearn-Merge and GA/NP improve the trade-off between forgetting and retention [30]. While similar metrics on retain sets can quantify preservation [30]-[34], [37], they risk masking subtle degradation in specialized domains.

Efficiency is another central concern, as full retraining is computationally prohibitive. Approximate unlearning demonstrates up to a tenfold speedup compared to retraining [27], [28], while federated and blockchain-enhanced approaches help mitigate single-point retraining costs [37]. Efficiency is typically measured in computational time and memory usage [27], [28], [40], though approximate methods often trade stronger guarantees for speed, and federated approaches introduce synchronization and reliability challenges.

Despite the breadth of existing approaches, no unified benchmark currently exists, leading to fragmented comparisons across studies. Most methods are validated only on small- or mid-scale models, raising concerns about scalability to frontier LLMs. Efficiency claims are often context-dependent, underscoring the need for robust and standardized evaluation pipelines.

### B. Towards Robust Unlearning for LLMs

Even when unlearning is successful, models remain vulnerable to adversarial recovery and unintended data leakage [15]. Studies have shown that Representation Misdirection for Unlearning (RMU), a leading unlearning method, is still susceptible to fine-tuning attacks. Deeb and Roger [51] showed that fine-tuning attacks can restore up to 88% of the pre-unlearning accuracy. Similarly, Łucki et al. [52] demonstrated that even minimal fine-tuning using as few as five forget-set samples or small amounts of unrelated retain data was sufficient to reinstate unlearned knowledge and behaviors using RMU.

Extending beyond RMU, recent studies on the reversibility of unlearning and the extraction of erased data have revealed a broader blind spot in most mainstream unlearning methods: the lack of robustness. Robustness is crucial to unlearning, as its real-world applicability depends on the model's ability to maintain security and resist adversarial attempts to recover forgotten data. Łucki et al. [52] experimented with two unlearning baselines RMU and Negative Preference Optimization (NPO) [53], and conducted five white-box adversarial attacks on unlearned LLMs, all of which successfully reversed the unlearning effects and recovered erased contents post-unlearning. Other studies have also highlighted the vulnerabilities of unlearning methods to MIA [32], jailbreaking attacks [54]-[56], and relearning attacks [43], [56], [57]. This has led to a new research direction on robustness in LLMs unlearning. Robust unlearning requires the unlearned data to remain forgotten, even under sophisticated adversarial attacks [15]. It should also avoid introducing new vulnerabilities to model security and privacy [58].

To meet these goals, most existing robust unlearning methods adopt adversarial training, which was originally developed to improve model robustness against adversarial inputs and is formulated as a saddle-point problem with inner and outer optimization loops [59]. In unlearning, adversarial training works as a two-player game: the defender updates the model to meet specific unlearning objectives, while the attacker generates adversarial samples or jail- breaking prompts to reverse the effects [60]. The Latent Adversarial Unlearning (LAU) framework [61] applies this training in the latent activation space and is compatible with gradient-based methods, where attackers seek perturbations that reintroduce forgotten data and defenders suppress them. Similarly, Latent Adversarial Training (LAT) [43] targets latent space attacks for specific competing tasks, such as relearning. The Random Mapping Adversarial Training (RMAT) framework [62] adopts meta-learning with representation corruption which maps sensitive data representations to noise responses such as "I don't know" and simulates recovery attempts during training to improve resistance against resurgence of forgotten information. Beyond adversarial training, circuit breaking leverages mechanistic interpretability to first localize neurons responsible for the target unwanted knowledge, then apply unlearning loss to these components [63]. This improves robustness against attacks such as paraphrase generalization, latent knowledge retention, relearning resistance, and side-effect avoidance, while preserving general capabilities. However, it is limited to factual edits and requires manual effort for localization, which in turn reduces scalability and efficiency.

*1) Technical Safeguards and Benchmarks for protecting unlearning LLMs:* Building on the need for robustness outlined above, recent research has introduced a range of technical safeguards to strengthen unlearning methods against recovery attacks. These safeguards span adversarial training and regularization, optimization-based loss functions, mechanistic localization, adversarial evaluation frameworks, model editing, and broader conceptual approaches. Table III summarizes these methods, their benchmarks, and key findings. Adversarial training remains the most widely explored safeguard, with variants such as FAST-BAT [64] and GradAlign [65] mitigating catastrophic overfitting and improving robustness against PGD adversaries on benchmarks like CIFAR and ImageNet [59], [66]. Extensions such as AdvUnlearn [60] further demonstrate resilience in diffusion models, strengthening robustness against prompt attacks while retaining utility. These methods significantly improve robustness but introduce runtime trade-offs and cannot fully prevent leakage under stronger adversarial conditions. Optimization-driven approaches offer complementary safeguards by modifying the training objective. RMU reduces hazardous knowledge while preserving utility on WMDP [67], while NPO [53] and RMAT [62] strike a balance between forgetting and retention, with RMAT mapping sensitive representations to noise-like responses to resist resurgence.

8TABLE II: Summary of existing studies on evaluating effective unlearning in LLMs.

| Category | Papers | Metric/Method | Key Findings |
|---|---|---|---|
| Evaluation: Effectiveness of Unlearning | [35], [36] [39], [40] | Forget set accuracy (LiRA, KL-divergence), Truth Ratio, ROUGE-L, Probability metrics | Methods like ICU/LiRA and SPUL show measurable forgetting; entropy-based bounds capture leakage more reliably than deterministic tests: teacher-student distillation (RKU) achieves high forget quality |
| Evaluation: Resistance to Membership Inference & Recovery Attacks | [38], [48] [42] | MIA success rate, Exposure metric, Utility-leakage auditing | DP-SGD and SISA retraining reduce MIA accuracy. ECO dynamically unlearns knowledge with minimal side-effects; auditing (ULIRA, LiRA-Forget) exposes residual memorization |
| Model Preservation: Utility Retention | [41], [47] [43] | Retain set accuracy, ROUGE-L, JSD, W.$\Delta$H, Harmonic mean utility score, General benchmarks (MMLU, AGIEval) | Structured unlearning (Split-Unlearn-Merge, GA/NP) retains general capabilities while forgetting targeted data; ECO and $DP^2$Unlearning preserve model utility better than baseline methods |
| Efficiency: Cost of Unlearning | [42], [50] | Update time, Computational cost, Memory overhead, Federated setup | Approximate unlearning $\sim$10x faster than retraining; federated/blockchain-enhanced frameworks avoid single-point retraining; competitions emphasize adaptive evaluation |
| Verification: Guarantees of Unlearning | [44], [49] | Post-hoc auditing KS-test, Truth ratio, Theoretical bounds | Auditing-based verification detects residual memorization. TruthfulQA and misconception-based tests reveal remaining hidden knowledge; theory-driven approaches ($DP^2$, certified unlearning) provide formal guarantees |

Despite promising results, these approaches are often sensitive to hyperparameter tuning and lack consistent generalization across benchmarks such as MMLU and WMDP.

Mechanistic unlearning methods aim to localize and erase knowledge at the circuit or neuron level. Manual localization outperforms automated deletion in both robustness and leakage reduction, showing that fine-grained tracing of internal representations enhances resilience [60], [63]. However, these methods remain constrained by scalability and efficiency, limiting their deployment on large-scale LLMs. Evaluation and defense frameworks such as DUA and RTT play a dual role in exposing vulnerabilities and proposing countermeasures. DUA reveals recovery of more than 50% of unlearned knowledge, while RTT recovers up to 88% of pre-unlearning accuracy [51], [56], [57]. In contrast, defensive adaptations such as LAU [61] improve forgetting by over 50% with minimal utility loss, though leakage remains detectable. These findings underscore both the progress made and the persistence of residual memorization. Model editing defenses (e.g., MEMIT, ROME) reduce factual recovery rates on benchmarks such as CounterFact and zsRE [52], [54], lowering attack success by up to 39.5 points. Yet, they remain vulnerable to indirect recovery pathways such as in-context reintroduction, limiting their robustness guarantees. At a broader level, surveys and conceptual critiques stress that exact unlearning is insufficient, as erased knowledge can often reappear under adaptive prompting or adversarial settings [15], [47], [55], [58].

Taken together, these safeguards provide partial defenses and important insights but fall short of delivering complete irreversibility. Recovery attacks such as RTT highlight the fragility of current methods, while the bi-level optimization of adversarial training and manual burden of mechanistic localization raise practical concerns. More efficient adversarial training variants such as free adversarial training [66] and fast adversarial training [64], [68] show promise by reducing training costs without severely compromising robustness. Integrating such methods into unlearning pipelines may pave the way towards scalable, efficient, and practically robust unlearning in large-scale LLMs.

### C. Promoting User Trust in LLMs

The rapid adoption of LLMs in applications that engage end-users through human-like conversations has brought user trust to the forefront of concern [69]. However, the inherent complexity of LLM architecture and their blackbox nature create significant challenges. End-users, regulators, and even developers often struggle to understand how outputs are generated, what data have been learned, to what extent the learned knowledge influence model behaviors, and also how learned data is handled throughout the model lifecycle [70], [71]. This opacity undermines confidence and raises concerns about safety, fairness, and accountability.

*1) Explainability and Interpretability:* One approach to foster user trust is enhancing transparency and interpretability, often framed within the movement of Explainable AI (XAI). XAI aims to provide mechanisms that reveal the reason behind model predictions and decision-making processes [72]-[74]. For example, Wang et al. [75] employed mechanistic interpretability to analyze circuits in GPT-2, identifying specific attention heads and their functional roles. Similarly, the Automatic Circuit Discovery (ACDC) algorithm [76] has been used to detect neurons associated with distinct model behaviors. Sarti et al. [77] introduced a toolkit for interpretability analysis of sequential generation in LLMs, focusing on internal representations and feature-importance scoring in transformer architectures.

XAI has become important in sensitive contexts such as healthcare, finance and law, where trust becomes critical: users must be confident in the validity and verifiability of LLMs' outputs, the overall trustworthiness of the system, and the minimization of uncertainty [78]-[80]. Chittimalla et al. [80] advanced this line of application in healthcare by proposing a comprehensive framework for assessing interpretability in LLMs applied to healthcare. Their framework integrates attribution-based methods such as Integrated Gradients [81] and SHapley Additive exPlanations (SHAP) [82] to explain model decisions in terms of feature relevance and importance. The framework provides critical insights on important features such as symptoms and medical history, which are closely aligned with expert annotations. Sun et al. [83] introduced a novel framework to generate text-based explanation for medical question-answering outputs using expectation-maximization (EM) inference approach to pinpoint and weight the most relevant evidence passages from medical textbooks. By focusing the model's attention on key textual evidence, the framework significantly improves explanatory quality, achieving between 6% to 9% absolute gains in ROUGE-1 and BLEU-4 scores on two medical explanation datasets (MQAE-diag and MQAE) compared to prior baselines.

TABLE III: Technical safeguards for protecting unlearning LLMs

| Category | Paper | Threat Model/Benchmark | Key Findings |
|---|---|---|---|
| Adversarial Training & Regularization | [64], [66] [60], [65] [59] | PGD adversaries, CIFAR-10/100, ImageNet, diffusion models | Improves robustness significantly while reducing training cost; FAST-BAT mitigates catastrophic overfitting; GradAlign prevents FGSM overfitting with runtime trade-off; AdvUnlearn strengthens diffusion models against prompt attacks while retaining utility |
| Loss-Function & Optimization Methods | [67], [68] [53], [62] [51] | WMDP, MMLU, hazardous knowledge removal | RMU reduces hazardous knowledge while preserving general utility; NPO balances forgetting with utility retention; RMAT lowers forget set accuracy to chance level while maintaining retain performance |
| Localized & Mechanistic Unlearning | [60], [63] | Adversarial relearning, probing | Manual mechanistic unlearning outperforms automated localization, showing stronger robustness and reduced leakage |
| Adversarial Evaluation & Defense Frameworks | [56], [61] [51], [57] | Recovery attacks on TOFU, WMDP, WHP; suffix-based recovery | DUA reveals >50% recovery of unlearned knowledge; LAU improves forgetting by >53% with minimal utility loss; RTT recovers up to 88% pre-unlearning accuracy, exposing leakage; WHP reduces surface familiarity but retains latent knowledge |
| Model Editing Defenses | [52], [54] | White-box/black-box factual recovery on CounterFact, zsRE | Max-Entropy reduces attack success by up to 39.5 points; model editing removes facts but remains vulnerable to recovery attacks |
| Systematic & Conceptual Approaches | [55], [58] [15], [47] | Broad review of security/ privacy risks; in-context reintroduction | Surveys provide taxonomy & metrics, and highlight mixed-privacy solutions; conceptual critiques show exact unlearning insufficient, as knowledge can be reintroduced in-context |

Additionally, Savage et al. [84] proposed using prompt engineering to imitate the clinician's clinical reasoning on LLMs, which offers an interpretable rationale for physicians to evaluate LLMs' responses and trustworthiness in medical diagnosis. A similar approach was also investigated by Qiu et al. [85] to explain medical visual questions answering through chain-of-evidence prompting, which guided models to generate step-by-step reasoning chains with explicit medical evidence for interpretability. Empirical results demonstrate that the chain-of-evidence approach significantly boosts interpretability and inference performance of medical visual question answering systems.

Collectively, these advances highlight how XAI methods can enhance trust in clinical applications by ensuring that the rationales are coherent, verifiable, and aligned with expert knowledge.

*2) Governance and Accountability:* Beyond technical interpretability and transparency, user trust in LLMs also requires regulatory accountability frameworks. A pioneering work in transparency governance is the Model Card system established by Mitchell et al. [86], which aims to provide standardized disclosure of data features and ethical disclosures for intended applications of trained machine learning models. This line of work was further expanded by other studies that applied to datasets for transparency enhancement [87], [88]. In the context of LLMs, these standardized documentation practices can secure accountability by ensuring that end-users and regulators can scrutinize models' pre-deployment.

Risk auditing and evaluation is another significant aspect in promoting user trust on LLMs applications. Weidinger et al. [89] proposed a taxonomy of six risk areas including (1) discrimination, hate speech and exclusion, (2) information hazards, (3) misinformation harms, (4) malicious uses, (5) human-computer interaction harms, and (6) environmental and socioeconomic harms for risk evaluation and emphasized independent evaluation as a trust-building measure. Similarly, Perez et al. [90] demonstrated the potential of red teaming as a systematic strategy to identify failure modes, biases, and harmful behaviors before models are released to end-users.

Critical sectors such as healthcare cannot yet rely on existing techniques for provable compliance with privacy regulations [2]. Governments and institutions are moving towards formalizing comprehensive AI governance standards. The European Union's General Data Protection Regulation (GDPR) [21] reinforces accountability by granting individuals rights over their personal data, including the right to be forgotten and the right to withdraw consent, thereby strengthening mechanisms for data governance and control. Additionally, the EU AI Act [22] establishes the first risk-based framework for AI regulation, prohibiting "unacceptable risk" applications such as biometric categorization and cognitive manipulation of vulnerable groups. Similarly, the NIST AI Risk Management Framework [23] provides a structured process for identifying, managing, and mitigating risks in AI deployment. Together, these initiatives embed accountability into law and policy, ensuring that user trust is supported not only by technical safeguards but also by institutional and regulatory oversight.

*3) Trust under Transparency Constraints:* While explainability and governance provide foundational mechanisms for fostering trust, full transparency in LLMs is rarely achievable due to model complexity and proprietary restrictions. As such, practical approaches focus on maintaining user confidence under constrained transparency, often through robustness testing, uncertainty communication, and system-level safeguards. Table IV summarizes these approaches. Failure detection and robustness testing have emerged as critical tools. Red-teaming strategies, including zero/few-shot probing, reinforcement learning, and adversarial prompt injection, systematically reveal harmful behaviors prior to deployment [90]. Complementary methods, such as chain-of-evidence pretraining, improve evidence prioritization in question answering, thereby balancing safety and accuracy [85]. Together, these approaches highlight the trade-offs between diversity and safety: highly constrained systems reduce harmful responses but may also limit creativity. Attribution-based frameworks, building on techniques discussed earlier, provide further accountability through attention visualization, Integrated Gradients, SHAP, and counterfactuals [80], [82], [83]. Recent work combines attribution with evidence-based inference (e.g., EM methods) to improve clinical NLP interpretability, though scalability remains a challenge in large models. Documentation standards, such as Model Cards and Datasheets, complement these methods by situating interpretability results within broader system disclosures [73], [74], yet they often remain descriptive rather than enforceable. Uncertainty communication directly targets user perception. User studies show that natural-language expressions of uncertainty reduce overreliance on model outputs [79], while diagnostic reasoning prompts improve interpretability in clinical contexts [84]. These methods demonstrate value but are context-dependent, with effectiveness shaped by user expertise and application domain. Finally, domain-specific risk analyses emphasize that healthcare remains uniquely sensitive. Beyond interpretability case studies, conceptual reviews warn of risks such as professional deskilling, self-referential feedback loops, and medicolegal liability in AI-assisted practice [29]. Mechanistic interpretability techniques, including circuit analysis and causal probing [76], [77], offer fine-grained insights into hidden model behaviours, though their computational intensity limits real-world deployment. Verification and validation frameworks are beginning to address these gaps, but robust scaling remains unresolved.

Taken together, these approaches illustrate that promoting trust under transparency constraints requires a layered strategy: combining robustness audits, attribution-based accountability, structured documentation, and uncertainty-aware communication, while also recognizing domain-specific risks. Bridging these practices with unlearning safeguards is an open challenge, as transparency must not only inform users but also ensure that erased knowledge cannot be recovered or reintroduced.

## IV. DISCUSSION

Building on the Literature Review, this section discusses findings across the three identified domains: unlearning evaluation and verification, robustness evaluation, and user trust in LLMs. The discussion addresses RQ1-RQ3, relates the findings to broader literature, maps them to technical solutions (Fig. 4), and highlights unresolved challenges. We also outline directions for integrated frameworks that can support reliable deployment of privacy- preserving AI.

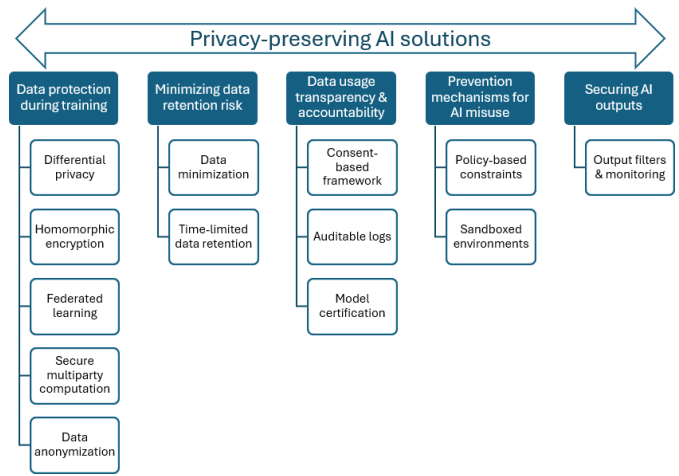

Fig. 4: Overview of solutions across privacy, unlearning, and trust.

Figure 4 summarizes existing approaches to privacy-preserving AI across the full lifecycle, spanning training, retention, accountability, misuse prevention, and secure outputs. These solutions form the baseline upon which more advanced safeguards, such as unlearning, must be built. While each category offers valuable protection, they also present limitations and remain vulnerable to compromise, underscoring the need for complementary techniques.

TABLE IV: Overview of approaches to enhancing user trust in LLMs under transparency constraints

| Category | Ref | Methods | Key Results |
|---|---|---|---|
| Failure Detection & Robustness | [85], [90] | Red-teaming (zero/few-shot, RL, BAD prompts); chain-of-evidence QA pretraining | Revealed harmful responses; highlighted diversity-safety trade-offs; improved evidence prioritization and SOTA accuracy. |
| Explainability via Attribution | [80], [82], [83], [89] | Attention visualization, IG, SHAP, counterfactuals, LIME; SHAP unification; EM inference for evidence | Improved interpretability and accountability; mitigated class imbalance; better QA and clinical NLP performance; scalability challenges remain. |
| Transparency & Documentation | [17], [73], [74] | Standardized Model Cards; narrative/multidisciplinary reviews (KDP, KR, RL link) | Enhanced transparency, fairness, and accountability; situated XAI in broader foundations. |
| Trust & Uncertainty Communication | [79], [84] | User studies on uncertainty (Qualtrics); diagnostic reasoning prompts on MedQA | Natural language uncertainty reduced overreliance; GPT-4 prompts increased interpretability of medical reasoning. |
| Domain-Specific Risks (Healthcare) | [29] | Conceptual/regulatory analysis of LLM use in healthcare | Identified risks of deskilling, self-referential loops, and medicolegal issues; emphasized need for transparency and human-AI collaboration. |
| Mechanistic Interpretability & Verification | [76], [77], [70], [75], [69] | Circuit analysis (ACDC, path patching, PCA/SVD, causal tracing); attribution toolkit (Inseq, CAT); V&V frameworks; forging maps with PoL logs | Recovered circuits; located biases and knowledge; identified induction heads; V&V improved trustworthiness; forging challenges unlearning but larger datasets aid validation. |

*1) Data Protection during Training:* Common safeguards include Differential Privacy (DP), Homomorphic Encryption (HE), Federated Learning (FL), Secure Multiparty Computation (SMPC), and Data Anonymization. DP introduces noise to training, reducing exposure of individual records, but cannot completely erase learned patterns once memorized. HE enables encrypted computation, preserving confidentiality, though at high computational cost that restricts its scalability to frontier LLMs. FL and SMPC reduce centralization risks by keeping data local or distributed, yet are vulnerable to poisoning attacks, adversarial updates, and heavy communication overhead. Data Anonymization, though widely used, has repeatedly been shown to fail against re-identification attacks when external linkages are available.

*2) Minimizing Data Retention Risks:* Minimization policies and time-limited retention restrict long-term exposure by ensuring sensitive data and residual artefacts are deleted after a set period. However, deletion at the data level does not necessarily remove influence embedded in trained model parameters. As such, even after retention policies are applied, models can still leak sensitive information through memorization or adversarial probing.

*3) Data Usage Transparency and Accountability:* Consent-based frameworks, auditable logs, and model certification schemes provide governance and oversight. These measures improve accountability and legal defensibility, allowing regulators and users to trace how data was collected and used. Yet they rely on compliance and enforcement rather than technical guarantees. Logs may themselves expose sensitive information if compromised, while certifications risk lagging behind rapid advances in AI development.

*4) Prevention Mechanisms for AI Misuse:* Policy-based constraints, sandboxed environments, and deny-by-default deployment rules restrict model misuse. These techniques improve containment and allow for structured monitoring. However, adversarial users can still exploit jailbreaks, prompt injections, or side channels to bypass constraints. Furthermore, tight restrictions may degrade utility by over-blocking benign outputs or limiting flexibility in downstream applications.

*5) Securing AI Outputs:* Output filters and monitoring systems aim to block the disclosure of personally identifiable information (PII), protected health information (PHI), or other sensitive content. These are increasingly integrated into inference gateways, particularly in regulated industries. Nonetheless, filter accuracy is a persistent challenge: false negatives leave room for leakage, while false positives increase latency and reduce usability.

*6) Limitations and Transition to Unlearning:* Together, these safeguards represent essential layers of defense. Yet they share a common drawback: none of them provide an explicit mechanism for *removing* the influence of data once it has been learned by a model. DP can reduce, but not eliminate, memorization; HE and FL prevent disclosure during training, but not post hoc recovery; retention policies and logs improve governance but cannot alter model parameters. Consequently, even well-protected systems remain vulnerable to adversarial probing, membership inference, and fine-tuning attacks that can resurface sensitive knowledge.

This gap motivates the need for machine unlearning, a complementary safeguard that directly targets the removal of unwanted data and its residual influence from model parameters. Unlike preventive or reactive measures, unlearning provides a technical path to enforce the "right to be forgotten" in LLMs, ensuring that erased data cannot be recovered through downstream attacks.

*7) Future Directions: Integrating Unlearning with Lifecycle Safeguards:* Future research should explore integrated frameworks where unlearning is combined with existing lifecycle protections. For example, coupling unlearning with auditable logs could provide verifiable deletion pipelines; combining unlearning with FL or SMPC could enforce forgetting in collaborative training settings; and linking unlearning with output monitoring could ensure that forgotten knowledge cannot reappear in generated content. Such hybrid approaches would bridge preventive, reactive, and active deletion strategies, providing stronger privacy guarantees and improving trust in real-world deployments of LLMs.

## V. Conclusion

This study has examined how machine unlearning can support privacy, security, and trust in LLMs by focusing on three central challenges.

First, we considered how forgetting is evaluated. Metrics such as forget set accuracy, truth ratios, and membership inference attacks provide partial evidence that unlearning has taken place. Yet these methods remain computationally demanding and lack the consistency required for standardized verification, leaving reliable assessment an open problem.

Second, we reviewed the resilience of unlearned models against adversarial recovery. Techniques including adversarial training, latent space defenses, and circuit based approaches improve robustness in some cases, but they impose significant overhead and are difficult to adopt widely, especially for smaller organizations relying on open source systems.

Third, we explored how trust can be supported when model complexity or proprietary constraints limit transparency. Technical safeguards alone are insufficient. Auditing pipelines, explainability methods, and regulatory frameworks such as GDPR and the EU AI Act are necessary complements if LLMs are to be used responsibly in sensitive settings.

Taken together, these findings show steady progress but no definitive solution. Robust and verifiable unlearning remains unresolved, and organizations must still weigh the benefits of LLMs against the risks of data retention. Moving forward will require more efficient deletion techniques, stronger defenses against adversarial recovery, and governance structures that reinforce accountability. Ultimately, accountable unlearning is not only a technical challenge but also a societal imperative, essential for building AI systems that protect both privacy and public trust.